\documentclass[preprint,12pt]{elsarticle}

\usepackage{amssymb}
\usepackage{amsmath}

\usepackage{algorithmic}
\usepackage{algorithm}
\usepackage{array}
\usepackage[caption=false,font=normalsize,labelfont=sf,textfont=sf]{subfig}
\usepackage{textcomp}
\usepackage{stfloats}
\usepackage{url}
\usepackage{verbatim}
\usepackage{graphicx}
\usepackage{booktabs}

\journal{ }

\begin{document}

\begin{frontmatter}



\title{Classifier-Free Diffusion-Based Weakly-Supervised Approach for Health Indicator Derivation in Rotating Machines: Advancing Early Fault Detection and Condition Monitoring}


\author[label1]{Wenyang Hu} 
\ead{hwy20@mails.tsinghua.edu.cn}
\affiliation[label1]{organization={Department of Mechanical Engineering},
 addressline={Tsinghua University}, 
 city={Haidian District},
 postcode={100084}, 
 state={Beijing},
 country={China}}

\author[label2]{Gaetan Frusque} 

\affiliation[label2]{organization={Intelligent Maintenance and Operations Systems},
 addressline={EPFL}, 
 city={Laussane},
 postcode={12309}, 
 country={Switzerland}}
\author[label1]{Tianyang Wang\corref{cor1}} 
\ead{wty19850925@mail.tsinghua.edu.cn}
 \cortext[cor1]{*=Corresponding author}

\author[label1]{Fulei Chu} 

\author[label2]{Olga Fink} 
\ead{olga.fink@epfl.ch}

\begin{abstract}
Deriving health indicators of rotating machines is crucial for their maintenance. However, this process is challenging for the prevalent adopted intelligent methods since they may take the whole data distributions, not only introducing noise interference but also lacking the explainability. To address these issues, we propose a diffusion-based weakly-supervised approach for deriving health indicators of rotating machines, enabling early fault detection and continuous monitoring of condition evolution. This approach relies on a classifier-free diffusion model trained using healthy samples and a few anomalies. This model generates healthy samples. and by comparing the differences between the original samples and the generated ones in the envelope spectrum, we construct an anomaly map that clearly identifies faults. Health indicators are then derived, which can explain the fault types and mitigate noise interference. Comparative studies on two cases demonstrate that the proposed method offers superior health monitoring effectiveness and robustness compared to baseline models.
\end{abstract}







\end{frontmatter}

\section{Introduction}
Intelligent health monitoring techniques for rotating machines are crucial for ensuring stable operation throughout their lifecycle. Health monitoring and early fault detection typically rely on data collected from accelerometer measurements, from which health indicators (HI) are learned or constructed to assess the machine's condition. These HIs can be viewed as a type of anomaly score that provides information on the severity of the anomalous condition \cite{hu2023wasserstein, wang2020extended, frusque2023semi}. By tracking the constructed HI, the evolution of the degradation severity in rotating machines can be monitored, enabling the optimization of effective maintenance strategies. One of the main challenges is distinguishing variations in the health indicator caused by changes in environmental and operational conditions or noisy environments from those that genuinely stem from abnormal conditions. The induced noise is particularly challenging as it is often complex, multi-sourced. 
 \par Rotating machines typically operate under healthy conditions for most of their lifecycle, which limits the availability of anomalous samples \cite{booyse2020deep}. To address these challenges, weakly-supervised or unsupervised health monitoring and early fault detection approaches have been developed\cite{rombach2024learning, frusque2023semi}. These models can be broadly categorized into classical machine learning-based methods \cite{heng2009rotating} and more recently developed deep learning methods\cite{fink2020potential}. Over the past decade, classical machine learning methods, such as Gaussian mixture models (GMM) and hidden Markov models(HMM), have been introduced to many rotating machine health monitoring and early fault detection tasks \cite{hong2019early}. Although these methods have achieved good results in some applications, they still lack the ability to construct robust and effective health indicators for complex time series data, such as the data with long time dependency or large noise interference \cite{zhao2019deep}. 
 \par In contrast, deep learning techniques have rapidly developed in recent years and have been successfully employed in various applications of rotating machines health monitoring\cite{fink2020potential}. 
 A widely used approach is to construct anomaly scores and monitor the health condition of assets with the reconstruction errors \cite{hsu2023comparison}. 
In reconstruction-based methods, autoencoder-based approaches, auto-regressive models, and methods mapping from operating conditions to sensor measurements are utilized \cite{li2019deep,lindley2021application,hsu2023comparison}. These methods aim to learn the distribution of healthy samples and use the deviations between the original and reconstructed samples to indicate the health condition of rotating machines. The underlying assumption is that the more anomalous a sample is, the greater its deviation from the healthy condition, resulting in a higher reconstruction error and a lower health indicator value. Although these methods have received significant attention, they often neglect the physical explanation for the constructed HI, leading to less reliable health monitoring results\cite{sun2022situ}. In addition, these methods may struggle to modeling some samples with complex distributions\cite{booyse2020deep}. 
 \par Recently, an increasing number of methods have adopted deep generative models, particularly Generative Adversarial Networks (GANs), due to their strong capabilities in capturing the statistical properties and temporal dynamics of time-series condition monitoring data\cite{yoon2019time,booyse2020deep,ren2023few}. These deep generative models can better reflect the distribution of healthy monitoring data, making anomalies easier to detect \cite{soleimani2022system,booyse2020deep}. 
 The concept behind these deep generative models is to learn the distribution of healthy data and utilize the distributional differences between generated samples and measured samples to construct HIs. These differences can be reflected not only in the reconstruction errors between the original data and the synthetic data but also in the scores given by the discriminator networks, which assign low scores to anomalous samples, or a combinations of both reconstruction errors and discriminator scores\cite{luleci2023cyclegan, li2024acwgan}. Although GAN-based health monitoring methods have proven effective in some applications, training a a GAN-based model remains challenging. In addition, the HI constructed with these methods often lack physical explanation. Moreover, many applications of this method attempt to model the distribution of the entire monitoring sample, including both potential fault features and external noise interference. This aim of modeling the whole sample's distribution may introduce extra noise, causing fluctuations in the HI curves. Therefore, constructing HIs based solely on the potential fault features is highly expected. 
 \par In this work, we propose a weakly supervised deep generative method called the Anomaly Map Guided Health Monitoring and Early Fault Detection method for rotating machines, based on the classifier-free diffusion model (AMG-CF-Diffusion). The classifier-free mechanism generates controlled healthy samples for the original condition monitoring data of each measured sample throughout the lifecycle of rotating machines. By comparing the generated samples with real-time monitoring data in the envelope spectral domain, an anomaly map is constructed. This anomaly map represents the differences between the generated samples and the original samples in the envelope spectral domain, revealing the fault characteristic frequency of the original samples. As a result, the HI for health monitoring can be constructed using the differences of the fault features rather than the whole sample distribution. This approach not only explains the fault types in rotating machines but also helps construct HIs directly from the differences in fault features. We evaluated the performance of our proposed method on two datasets, comparing its robustness to noise interference and its effectiveness in health monitoring and early fault detection with other state-of-the-art (SOTA) methods. The results show that our proposed method is more robust to noise interference under different signal-to-noise-ratios (SNR). Additionally, our results demonstrate that AMG-CF-Diffusion can monitor the health condition of rotating machines and detect early faults more effectively.

\section{Preliminaries}

\subsection{Denoising Diffusion Probabilistic Models}
\par The denoising diffusion probabilistic models (DDPM) were first proposed by Ho et al. in 2020\cite{ho2020denoising}. After this method was proposed, it has achieved a series of applications, such as data augmentation, image generation, and video generation\cite{wang2024data,epstein2023diffusion,ho2022video}. The DDPM generally consists of two processes: the forward process and the backward process. In the forward process, Gaussian noise is progressively added to the original training sample $\boldsymbol{x}_{0}$ over $T$ time steps. This forward process $q$ at step $t$ can be expressed as:
\begin{equation}
 q(\boldsymbol{x}_{t}|\boldsymbol{x}_{t-1}) = N(\boldsymbol{x}_{t}; \sqrt{\alpha_t}\boldsymbol{x}_{t-1},(1-\alpha_t)\boldsymbol{I}) 
 \label{eq4}
\end{equation}
where $t \in \{ 1, ..., T\}$. $1-\alpha_t$ corresponds to the noise intensity at each step.
For an arbitrary step $t$, the processed sample $\boldsymbol{x}_t$ can be expressed as:
\begin{equation}
 \boldsymbol{x}_{t} = \sqrt{\overline{\alpha}_t}\boldsymbol{x}_{0} + \sqrt{1-\overline{\alpha}_t}\boldsymbol{\zeta}, \boldsymbol{\zeta} \sim N(0,\boldsymbol{I})
 \label{eq5}
\end{equation}
where $\overline{\alpha}_t =\prod_{i=1}^{t} \alpha_i$. 

Conversely, the reverse process aims to denoise and recover the initial distribution from the noisy samples $\boldsymbol{x}_{T}$. For an arbitrary time step $t$, the reverse denoising process $p$ can be expressed as:
\begin{equation}
 p_{\theta}(\boldsymbol{x}_{t-1}|\boldsymbol{x}_{t}) = N(\boldsymbol{x}_{t-1};\mu_{\theta}(\boldsymbol{x}_{t},t), \Sigma_{\theta}(\boldsymbol{x}_t,t)) 
\end{equation}
This process can be modeled using a neural network parameterized by $\theta$. The goal of training this network is to approximate the posterior distribution $q(x_{t-1}|x_t,x_0)$ by minimizing the following loss function:
\begin{equation}
 \min_{\theta} \Vert \boldsymbol{\zeta}-\boldsymbol{\epsilon}_\theta(\sqrt{\overline{\alpha}_t} \boldsymbol{x}_{0}+\sqrt{1-\overline{\alpha}_t}\boldsymbol{\zeta},t) \Vert, \boldsymbol{\zeta} \sim N(0, \boldsymbol{I})
\end{equation}
where $\epsilon_\theta$ represents the output of the network parameterized by $\theta$. 

Once the training process is completed, samples can be synthesized as:
\begin{equation}
 \boldsymbol{x}_{t-1,0} = \frac{1}{\sqrt{\alpha_t}}(\boldsymbol{x}_{t,0} - \frac{1-\alpha_t}{\sqrt{1-\overline{\alpha}_t}}\boldsymbol{\epsilon}_\theta(\boldsymbol{x}_{t,0},t)) + \sigma_t \boldsymbol{\zeta}, t \in \{ 1, 2,..., T\}
 \label{eq8}
\end{equation}
where $\sigma_t = \sqrt{1-\alpha_t}$.
\subsection{Controlled samples generation based on Diffusion models}
\par Controlled sample generation refers to the process of systematically generating sample data or instances under specific constraints and conditions. In practical applications, many of these tasks require label information. By incorporating the classifier guidance mechanism, the diffusion model can generate samples based on class labels\cite{mukhopadhyay2023diffusion}. This classifier guidance mechanism can be expressed as:
\begin{equation}
 \nabla\log p(\boldsymbol{x}_t|y) = \nabla\log p(\boldsymbol{x}_t) +\nabla \log p(y|\boldsymbol{x}_t)
 \label{eq_cg}
\end{equation}
where y refers to the label information and $p(\boldsymbol{x}_t|y)$ is a classifier. Here, $\nabla\log p(\boldsymbol{x}_t|y)$ represents the gradient of the controlled synthesizing process corresponding to the related label information $y$. It can be seen that $\nabla\log p(\boldsymbol{x}_t)$ is the gradient for the unconditional synthesizing process, and $\nabla\log p(y|\boldsymbol{x}_t)$ corresponds to the classifier gradient.
\par However, classifier-guidance Diffusion models (CG-Diffusion) require an additional well-trained classifier to guide the sample synthesis, as reflected in Eq. (\ref{eq_cg}). During inference with the trained Diffusion model under the guidance of the trained classifier, additional gradient calculations are needed at each step. Moreover, the quality of the generated controlled samples depends on the classifier's performance. If the training data contains significant noise, the distribution of noisy samples $\boldsymbol{x}_T$ after the forward process can be similar across different sample classes, complicating the training of a robust classifier. 
\par To address these issues, the classifier-free mechanism has been proposed\cite{ho2022classifier}. The core principle of this mechanism is to replace the explicit classifier and its gradient with an implicit classifier. According to Bayes' theorem, the gradient of the classifier can be expressed in terms of conditional and unconditional generation probabilities:
\begin{equation}
 \nabla_{\boldsymbol{x}_t} \log p(y|\boldsymbol{x}_t) = \nabla\log p(\boldsymbol{x}_t|y) - \nabla_{\boldsymbol{x}_t} \log p(\boldsymbol{x}_t)
 \label{eq_cf}
\end{equation}
 By substituting the classifier gradient $\nabla_{\boldsymbol{x}_t} \log p(y|\boldsymbol{x}_t)$ into Eq. (\ref{eq_cg}), the classifier-free mechanism can be derived as:
 \begin{equation}
 \overline{\boldsymbol{\epsilon}}_{\theta}(\boldsymbol{x}_t,t,y)=(w+1)\boldsymbol{\epsilon}_{\theta}(\boldsymbol{x}_t,t,y) - w\boldsymbol{\epsilon}_{\theta}(\boldsymbol{x}_t,t)
 \label{eq_cf_2}
 \end{equation} 
Based on classifier-free Diffusion models (CF-Diffusion), various samples can be synthesized conditionally according to Eq. (\ref{eq_cf_2}). When using a well-trained model for inference, the generation effect can be adjusted through the guidance coefficient $w$ to balance the fidelity and diversity of the generated samples. 

\section{Methodology}
\par 
In this paper, we propose an anomaly map-guided health monitoring and early fault detection method based on a CF-Diffusion model.
The workflow of our proposed method is illustrated in Fig. 1. In the first step, we use healthy samples and minor anomalies to train a specifically designed classifier-free Diffusion model. With this well-trained end-to-end model, we input each measured sample of original condition monitoring data to generate samples that have a distribution similar to that of the original healthy samples. Subsequently, we apply the Hilbert transform to both the generated samples and the original condition monitoring samples to obtain an anomaly map in the envelope spectral domain. This anomaly map effectively indicates the type of fault present in the rotating machine. In addition, this anomaly map can be used to construct a HI corresponding to potential faults. 

\begin{figure}[!t]
\centering
\includegraphics[width=2.7in]{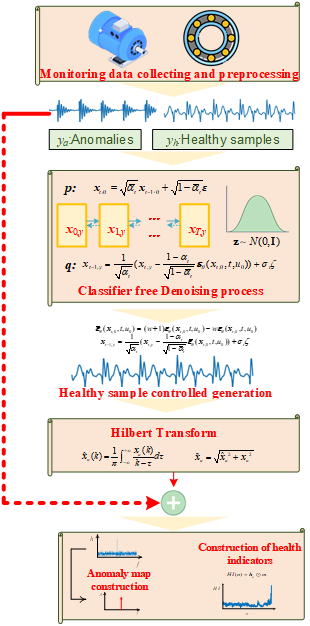}
\caption{Working flow of the proposed AMG-CF-Diffusion model for health management of rotating machines}
\label{fig_1}
\end{figure}

\subsection{Condition monitoring data collection and preprocessing}
\par The initial step involves collecting raw condition monitoring data using the the collection system. Subsequently, this raw data undergoes preprocessing. Specifically, the window cropping method is employed to segment the raw condition monitoring data into multiple samples. This approach not only reduces computational time but also ensures the diversity of the training samples. The resulting preprocessed data is denoted as $\boldsymbol{x}_0$.

\subsection{Training of conditional Diffusion model with classifier-free guidance}
\par In this section, we aim to develop a CF-Diffusion model tailored for condition monitoring of rotating machinery. This model is trained using preprocessed condition monitoring data $\boldsymbol{x}_0$, comprising healthy samples, $\boldsymbol{x}_h$, and anomalies, $\boldsymbol{x}_a$. The labels for these samples are denoted as $y_h$ and $y_a$, respectively. The training process involves two main sub-processes. 

\begin{enumerate}
 \item \textbf{Forward Process:} Noise is progressively added to the original training data, described by the equation:
\begin{equation}
 \boldsymbol{x}_{t,0} = \sqrt{\overline{\alpha}_t}\boldsymbol{x}_{0} + \sqrt{1-\overline{\alpha}_t}\boldsymbol{\zeta}, \boldsymbol{\zeta} \sim N(0,\boldsymbol{I}), t \in \{ 1, 2,..., T\}
 \label{eq11}
\end{equation} 
Here, $T$ represents the total number of time steps in the forward process. 
\item \textbf{Reverse Process:} This sub-process aims to remove the noise added during the forward process and recover the original distribution. It is mathematically represented as:
\begin{equation}
 \boldsymbol{x}_{t-1,0} = \frac{1}{\sqrt{\alpha_t}}(\boldsymbol{x}_{t,0} - \frac{1-\alpha_t}{\sqrt{1-\overline{\alpha}_t}}\boldsymbol{\epsilon}_\theta(\boldsymbol{x}_{t,0},t,u_0)) + \sigma_t \boldsymbol{\zeta}
 \label{eq12}
\end{equation}
where 
\begin{equation}
	u_0 = y_0c, c=\left\{
	\begin{aligned}
		1 & , & \rho \leq uncon\\
		0 & , & \rho > uncon
	\end{aligned},
 \rho \sim U(0,1)
	\right.
 \label{eq_}
\end{equation}
The parameter $uncon$ is a hyper-parameter that controls the mix of unconditional and conditional models during the joint training phase. The $\boldsymbol{\epsilon}_\theta$ represents the neural network parameterized by $\theta$, tailored to this model. 

\end{enumerate}
The training objective is to minimize the deviation of the noise model from the desired trajectory, formalized as:
\begin{equation}
 \min_{\theta} \Vert \boldsymbol{\zeta}-\boldsymbol{\epsilon}_\theta(\sqrt{\overline{\alpha}_t} \boldsymbol{x}_{0}+\sqrt{1-\overline{\alpha}_t}\boldsymbol{\zeta},t,u_0) \Vert,
 \label{eq_m_goal}
\end{equation}



\subsection{Generating Healthy Samples for Anomaly Detection}
\par Once the network converges, it can be utilized to generate samples that closely mimic the distribution of the initial healthy samples. These synthetic samples are crucial for generating anomaly maps by highlighting discrepancies between the original input and the synthetic samples. In the first step, anomaly samples, denoted as $\boldsymbol{x}_a$, are fed into the network. Here, the input labels are assigned as healthy, $y_h$, to ensure that the synthetic samples resemble the distribution of healthy condition monitoring samples. The synthesizing process is defined by the equation:
\begin{equation}
 \boldsymbol{x}_{t-1,a} = \frac{1}{\sqrt{\alpha_t}}(\boldsymbol{x}_{t,a} - \frac{1-\alpha_t}{\sqrt{1-\overline{\alpha}_t}}\boldsymbol{\overline{\epsilon}}_\theta(\boldsymbol{x}_{t,a},t,u_h)) + \sigma_t \boldsymbol{\zeta} 
 \label{eq_sampling}
\end{equation}
 Here $N$ is the total number of measured samples and $T$ represents the total number of time steps of forward process. $\overline{\boldsymbol{\epsilon}}_\theta$ is a linear combination of the conditional and unconditional model outputs, similarly combined as in Eq. (\ref{eq_cf_2}). The resulting synthetic samples are denoted as $\boldsymbol{x}_{h,a}$. 
 
\subsection{Anomaly map construction}
\par After obtaining synthetic samples, anomaly maps are constructed in the envelope spectral domain. First, the Hilbert transform is applied to the anomalous sample $\boldsymbol{x}_a$ and its corresponding synthetic sample $\boldsymbol{x}_{h,a}$. The transform results, as shown in Eq. (\ref{eq_hil}), are used to obtain the analytic signals, which are represented in Eq. (\ref{eq_ana}). Based on these analytic signals, their spectrum is obtained through Fast Fourier Transform (FFT). This spectrum of analytic signals is regarded as the envelope spectrum of the original signals, denoted as $\boldsymbol{h}_a$ and $\boldsymbol{h}_{h,a}$, respectively. In this context, fault features can be identified by subtracting the two envelope spectra, as shown in Eq. (\ref{eq_ano_map}).
\begin{equation}
 \hat{\boldsymbol{x}}_a(k) = \frac{1}{\pi}\int_{-\infty}^{\infty}\frac{\boldsymbol{x}_a(\tau)}{k-\tau}d\tau,\quad
 \hat{\boldsymbol{x}}_{h,a}(k) = \frac{1}{\pi}\int_{-\infty}^{\infty}\frac{\boldsymbol{x}_{h,a}(\tau)}{k-\tau}d\tau 
 \label{eq_hil}
\end{equation}
\begin{equation}
 \Tilde{\boldsymbol{x}}_a = \sqrt{\hat{\boldsymbol{x}}_a^2+\boldsymbol{x}_a^2},\quad 
 \Tilde{\boldsymbol{x}}_{h,a} = \sqrt{\hat{\boldsymbol{x}}_{h,a}^2+\boldsymbol{x}_{h,a}^2}
 \label{eq_ana}
\end{equation}
\begin{equation}
 \boldsymbol{A}=[\overbrace{0,...}^{p-1},1,...,0],\quad p=\mathop{\arg\max}\limits_{l} |\boldsymbol{h}_a(l)-\boldsymbol{h}_{h,a}(l)|
 \label{eq_ano_map}
\end{equation}
An anomaly map $\boldsymbol{A}$ is essentially a pulse vector consisting of only 0s and 1s, where the index corresponding to the pulse location indicates the anomaly frequency. This anomaly frequency can be compared with the theoretical fault feature frequency to determine kind of potential faults the rotating machines have. This anomaly map will be used to furthermore construct HIs for health monitoring purpose. 
\subsection{HI construction}
\par After generating the anomaly maps for specific anomalies, these maps are used to construct HIs for each measured sample of the monitoring data. This method differs from traditional approaches that construct anomaly scores by comparing distributions between reconstructed samples. Instead, we utilize the detailed fault feature information pinpointed by the anomaly maps. Based on these anomaly maps, the differences in distribution between fault features are directly used to construct health indicators, as shown in Eq. (\ref{eq_HI}).
\begin{equation}
     HI(n) = |\boldsymbol{h}_a(l)-\boldsymbol{h}_{h, a}(l)|\boldsymbol{A}^T 
 \label{eq_HI}
\end{equation}
Here, $n$ represents the index of the condition monitoring data measured sample during the entire life cycle of the rotating machinery. 
\par Utilizing the constructed HI, we can effectively undertake various health management tasks such as degradation tracking and continuous health monitoring. Furthermore, the HIs, when combined with appropriate threshold selection strategies, enable the early detection of faults. This proactive approach allows for timely interventions, enhancing the reliability and efficiency of the monitored systems.

\section{Experiments}
\par In this section, we present two case studies to demonstrate the effectiveness and the robustness of our proposed method. The first case involves a simulation experiment, which allows direct access to the ground truth HI information and provides full interpretability of the results, which can be more challenging to achieve with real faults. Additionally, a real experimental data will also be adopted to further illustrate the superiority of the proposed method.
\subsection{Case I: Bearing simulation}
\begin{figure}[!t]
\centering
\includegraphics[width=3.0in]{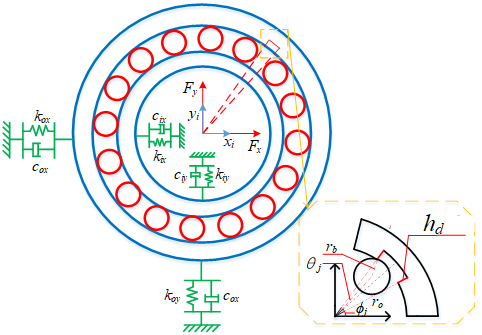}
\caption{The outer-race defect simulation model for a roller-bearing}
\label{fig_2}
\end{figure}
\subsubsection{Experimental setting}
\par In this case study, we used a simulation dataset specifically designed for conducting comparative studies and demonstrating the superior performance and robustness of our proposed method. This dataset was generated using a simulation model of a roller bearing with an outer-race fault, as depicted in Fig. \ref{fig_2}\cite{cui2017quantitative}. It captures the steady-state response of the simulation model in the vertical direction across different defect depths. The fundamental simulation parameters are detailed in Table \ref{Table_simu}. It is important to note that for the purposes of this experiment, states exhibiting minimal impairment were classified healthy. In these states, the defect depth $h_d$ was set to $1 \times 10^{-3}$ mm. States indicating more significant damage, where the defect depth was $100 \times 10^{-6}$ mm, contributed a smaller number of anomalies.Additionally, the experiment included four distinct fault conditions with varying defect depths $h_d$, as specified in Table \ref{Table_simu_ex}. This setup allowed for a thorough evaluation of the model's effectiveness across a spectrum of fault severities.
\begin{table}[htbp]
\centering
\caption{Simulation parameters for the simulation model}
\begin{tabular}{cc}
\toprule
Geometric parameters & Values \\
\midrule
Roller diameter/mm & 8.4 \\
Pitch diameter/mm & 71.5 \\
Number of rollers & 16 \\
Contact angle/($^\circ$) & 15.17 \\
Rotating frequency/(Hz) & 33.33\\
Sampling frequency/(Hz) & 20000 \\
sample length & 512\\
number of samples for each experimental setting & 400\\
\bottomrule
\label{Table_simu}
\end{tabular}

\end{table}

\begin{table}[htbp]
\centering
\caption{experimental setting}
\begin{tabular}{cccc}
\toprule
index & Defect depth $\boldsymbol{h}_d$($\times 10^{-6}$mm) & conditions & number\\
\midrule
1 & 0.001 & Healthy & 400 \\
2 & 0.01 & damage level 1 & 400 \\
3 & 0.1 & damage level 2 & 400 \\
4 & 0.5 & damage level 3 & 400 \\
5 & 1.0 & damage level 4 & 400 \\
6 & 100 & Anomaly & 10 \\
\bottomrule
\label{Table_simu_ex}
\end{tabular}
\end{table}

\begin{figure}[!t]
\centering
\includegraphics[width=2.7in]{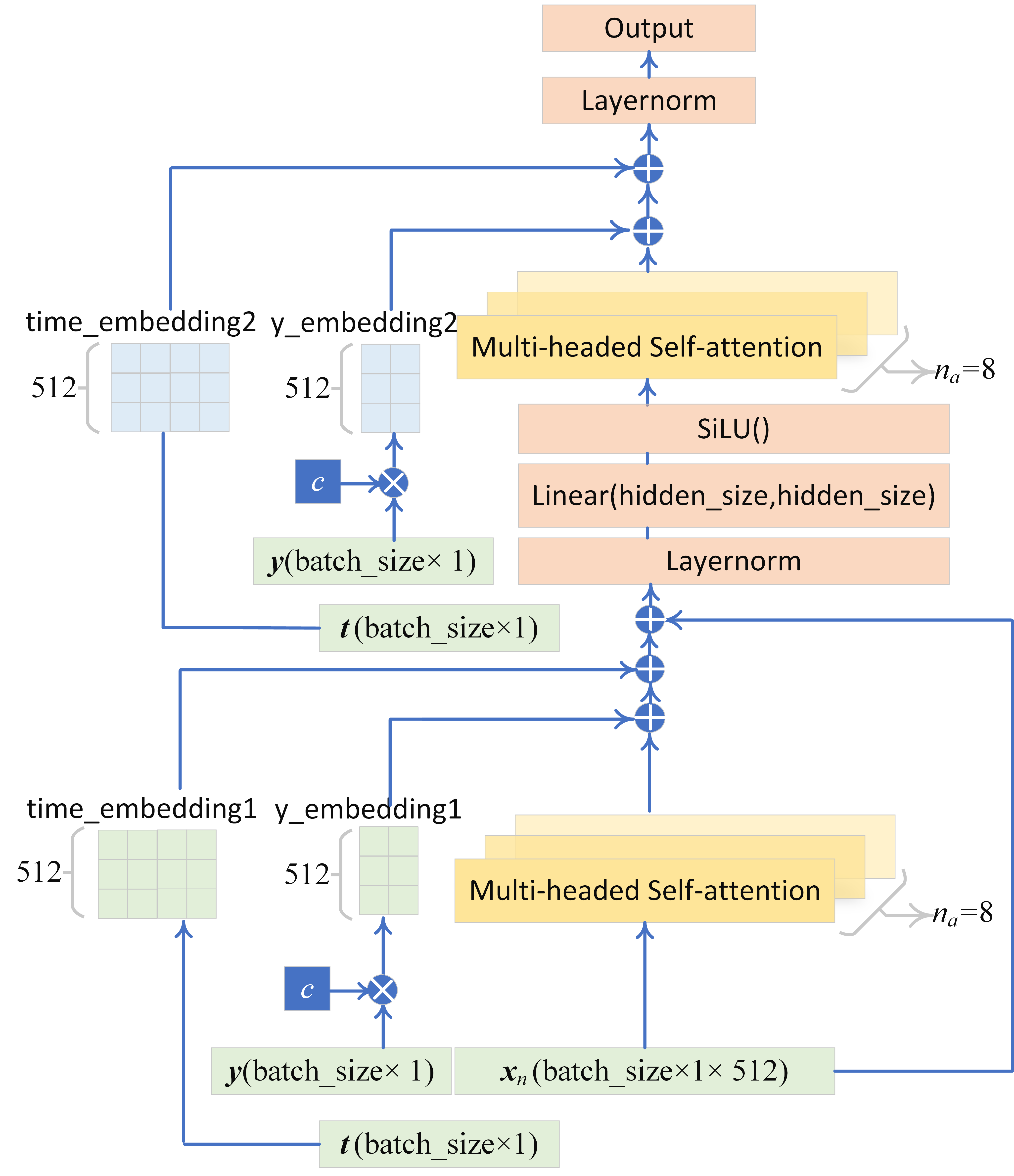}
\caption{The Configuration of the AMG-CF-Diffusion model for health
management in rotating machinery Systems}
\label{fig_nnset_1}
\end{figure}

\par First, the network of the CF-Diffusion model was trained using both anomalous and healthy samples. The 
architecture of this network is illustrated in Fig. \ref{fig_nnset_1}, which highlights the integration of two multi-headed self-attention layers. The dimension of the hidden layers was matched to the size of the input samples, set at 512. In addition, the samples' label information $y$ and diffusion time steps $t$ were embedded into the model. To optimize the network, the Adam optimization algorithm was employed aiming to achieve the goal specified in Eq. \ref{eq_m_goal}. The learning rate was set at 0.001, with a batch size of 16. 

\par After training the CF-Diffusion model, we proceeded with synthesis process to generate controlled healthy samples in accordance with Eq. (\ref{eq_sampling}). For this, the classifier-free parameter $w$ was set at 3, and the unconstrained parameter $uncon$ was set at 0.1. The ordinary form of the original healthy samples, anomalies, and the synthesized samples across time domain, spectral domain, and envelope spectral domain are displayed in Fig. \ref{fig_od_form_simu}. Qualitative analysis from Fig. \ref{fig_od_form_simu} suggests that the samples controlled and generated for the abnormal conditions closely mimic the distribution of the original healthy samples. This is evident from their spectra, which feature two spectral peak clusters as highlighted by the dotted boxes. Subsequent to this process, an anomaly map $\boldsymbol{A}$ was generated as per Eq. (\ref{eq_ano_map}), and can be viewedin Fig. \ref{fig_od_form_simu2}. By comparing this anomaly map with the theoretical fault feature frequency distribution typical of rotating machines, we can deduce the potential fault type present in this anomaly. In this instance, the anomaly map reveals a pulse near the outer-race fault feature frequency, confirming the same outer-ring fault type as that of the anomalies. Therefore, this precisely constructed anomaly map can be effectively utilized to diagnose potential faults in rotating machines.
\subsubsection{Results and discussions}
\begin{sloppypar}
\par Afterward, HIs for various experimental settings listed in TABLE \ref{Table_simu_ex} were constructed using the derived anomaly map to monitor the degradation process of the bearings, as outlined in Eq. (\ref{eq_HI}). The results are illustrated in Fig. \ref{fig_comp_simu}(a). To further evaluate the robustness of our proposed method, we introduced noise to the original experimental data to create datasets with varying signal-to-noise ratios (SNR). We then compared our method against established unsupervised or weakly-supervised baseline models for time series anomaly detection, including Feature Encoding with Auto-encoders for Weakly-supervised Anomaly Detection (FEAWAD)\cite{zhou2021feature}, the anomaly-transformer\cite{xu2021anomaly}, and the Wasserstein generative digital twin (WGDT)\cite{hu2023wasserstein}. The comparative results are displayed in Fig. \ref{fig_comp_simu} (b)-(d). Despite strong noise interference at -20dB, the HIs generated by our method consistently reflected the degradation trend of outer ring fault damage, whereas the HI constructed from FEAWAD failed to accurately track the degradation trend at noise levels above -10dB. Similarly, the Anomaly-transformer and WGDT models struggled to trace the bearing degradation trend accurately under high noise conditions exceeding -10dB. These findings quantitatively demonstrate the superior robustness of our proposed method.
\end{sloppypar}
\par Furthermore, we employed quantitative metrics such as cosine similarity and the Pearson correlation coefficient, detailed in Eq. \ref{metric_simu1} and Eq. \ref{metric_simu2}, to cassess the alignment between the actual bearing outer-ring defect depth and the HIs constructed by different models under various noise conditions. The results, depicted in Fig. \ref{fig_comp_simu_2}, show that as noise levels increased, the HIs from our method consistently maintained high similarity with the actual degradation levels of the bearing, indicating a superior ability to accurately track real degradation. In contrast, the similarity indices HIs from the other baseline models exhibited decreasing or fluctuating trend with increasing noise levels, underscoring the enhanced robustness of our approach against strong noise interference.
\subsubsection{Ablation studies}
\par We conducted two ablation experiments to validate the effectiveness of the proposed methods. These experiments primarily assessed the impact of incorporating the anomaly map in the envelope domain, comparing it against the performance of traditional reconstruction error based methods. The experimental approach involved comparing the HIs constructed using the reconstruction-errors of both the original samples and synthetic samples generated by the CF-Diffusion model across time domain and envelope spectrum, against our proposed Anomaly map guidance mechanism based on the CF-Diffusion model. The results of these experiments are illustrated in Fig. \ref{fig_simu_ablation}. They clearly show that the HI created using the anomaly map guidance mechanism exhibit a higher correlation with the actual degradation of the bearing, particularly as noise levels increase. This indicates that incorporating anomaly map guidance mechanism significantly enhances the model's ability to accurately track degradation, underscoring its superiority over traditional methods. 
\begin{equation}
 cos(\boldsymbol{HI}, \boldsymbol{h}_d) = \frac{\boldsymbol{HI} \cdot \boldsymbol{h}_d}{\|\boldsymbol{HI}\| \|\boldsymbol{h}_d\|}
 \label{metric_simu1}
\end{equation}
\begin{equation}
 r(\boldsymbol{HI},\boldsymbol{h}_d) = \frac{(\boldsymbol{HI} - \bar{\boldsymbol{HI}})^T (\boldsymbol{h}_d - \bar{\boldsymbol{h}_d})}{\| \boldsymbol{HI} - \bar{\boldsymbol{HI}} \| \| \boldsymbol{h}_d - \bar{\boldsymbol{h}_d} \|}
 \label{metric_simu2}
\end{equation}

\begin{figure}[!t]
\centering
\includegraphics[width=3.5in]{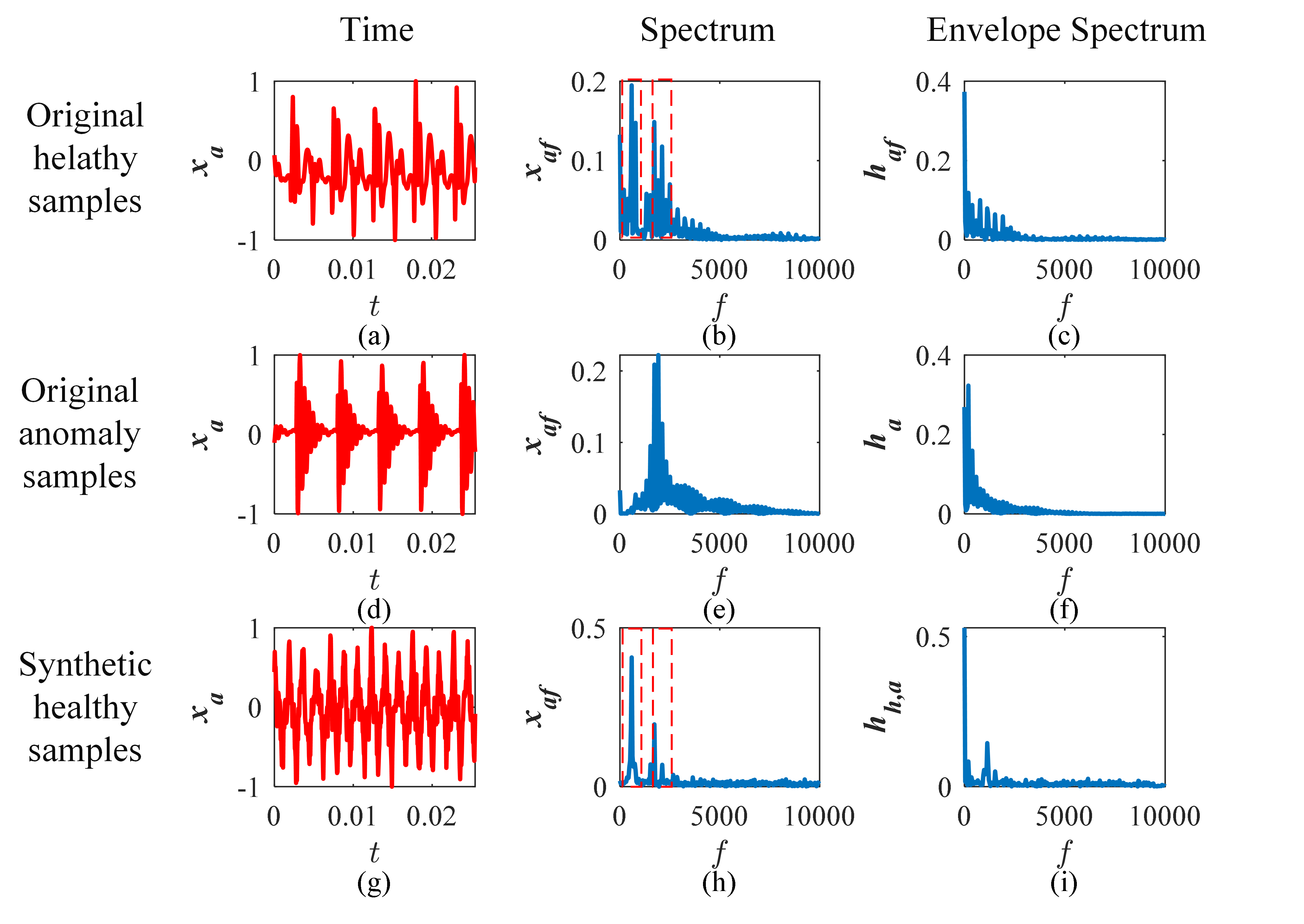}
\caption{The ordinary form of the original healthy samples, anomalies, and the synthesized samples in time domain, spectral domain, and envelope spectral domain.}
\label{fig_od_form_simu}
\end{figure}

\begin{figure}[!t]
\centering
\includegraphics[width=3.5in]{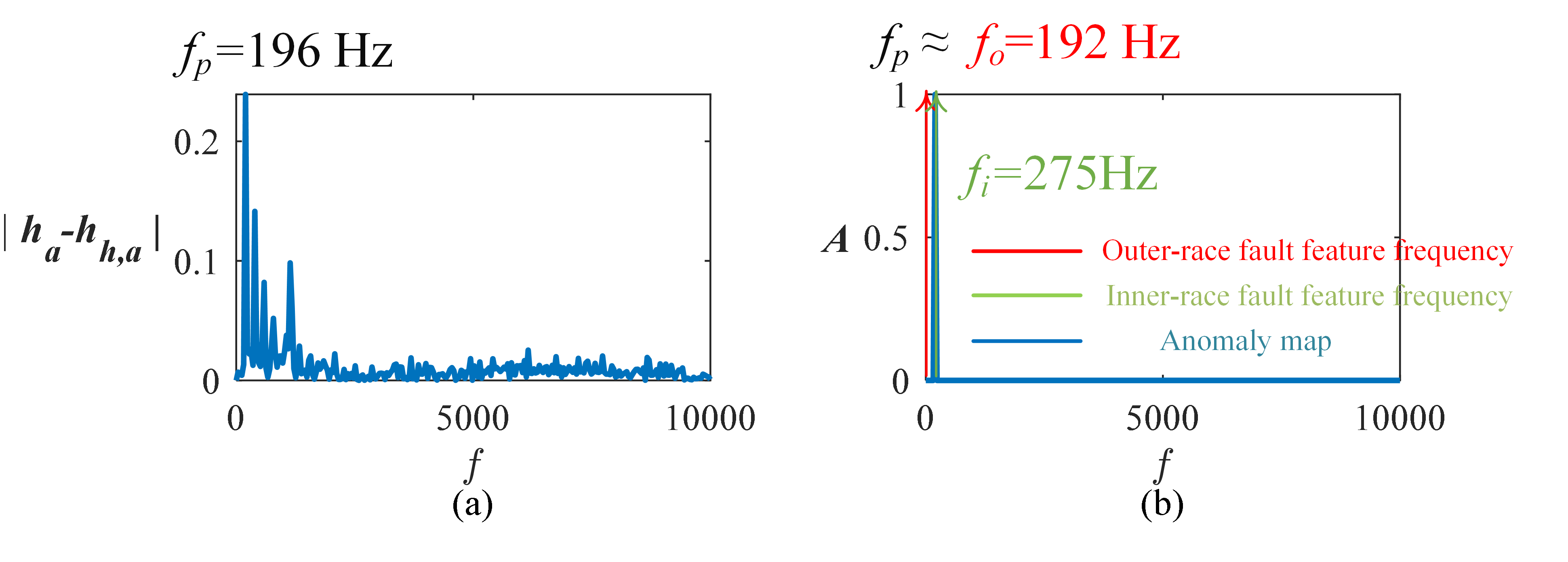}
\caption{The difference between the envelope spectra of synthetic healthy samples and original abnormal samples the obtained anomaly map. (a) The absolute value difference between the generated sample and the original sample $|\boldsymbol{h}_a-\boldsymbol{h}_{h,a}|$ (b) The obatined anomaly map $A$ }
\label{fig_od_form_simu2}
\end{figure}

\begin{figure}[!t]
\centering
\includegraphics[width=3.5in]{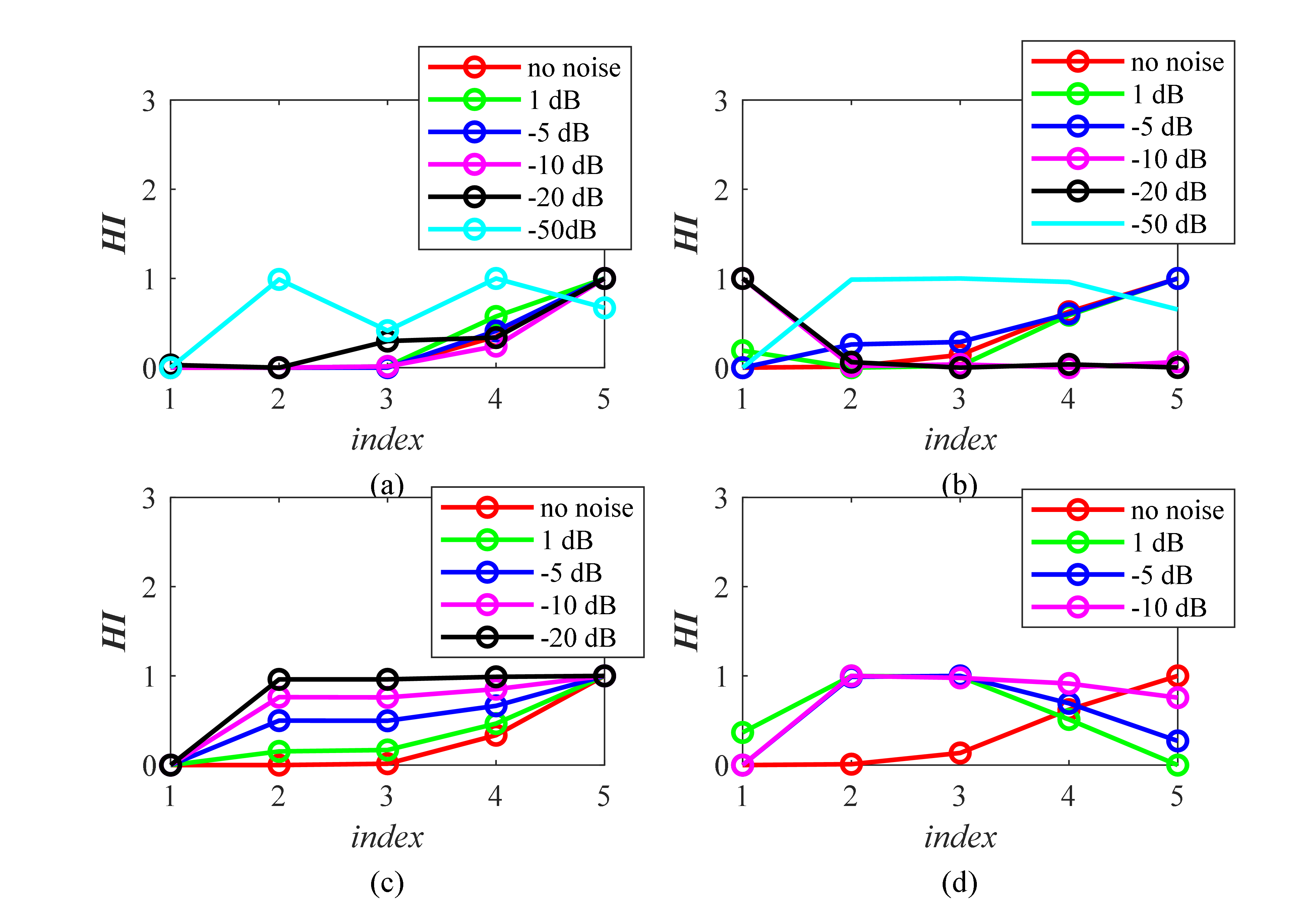}
\caption{The HI curves constructed with our proposed method and different baseline models under different noise interferences. (a) Anomaly map guidance mechanism based on CF-Diffusion (b) FEAWAD (c) Anomaly-Transformer (d) WGDT}
\label{fig_comp_simu}
\end{figure}

\begin{figure}[!t]
\centering
\includegraphics[width=3.5in]{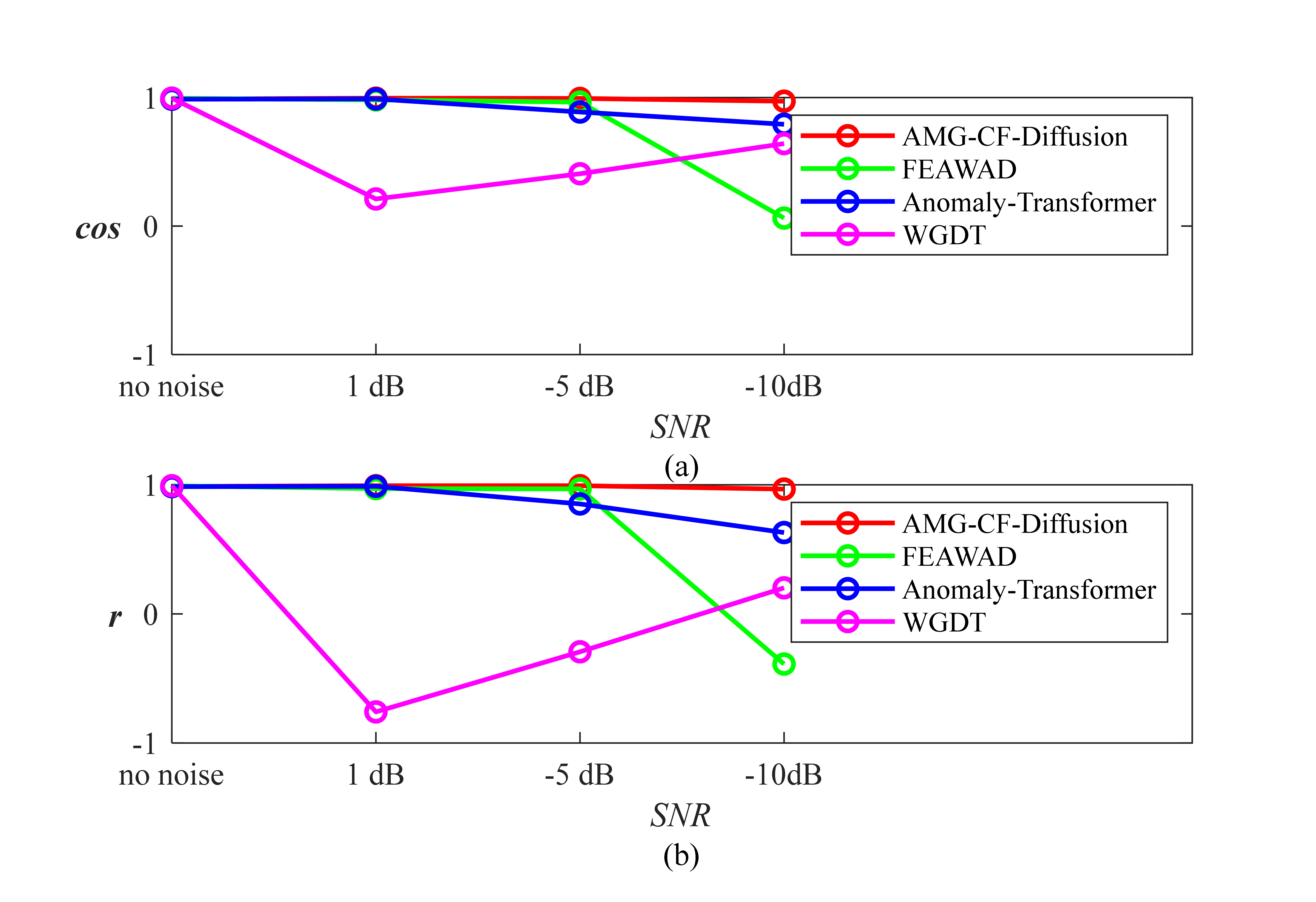}
\caption{The comparative analysis of the similarity between Actual degradation and HI curves generated using various baseline models. (a) cosine similarity (b) Pearson Correlation Coefficient}
\label{fig_comp_simu_2}
\end{figure}

\begin{figure}[!t]
\centering
\includegraphics[width=3.5in]{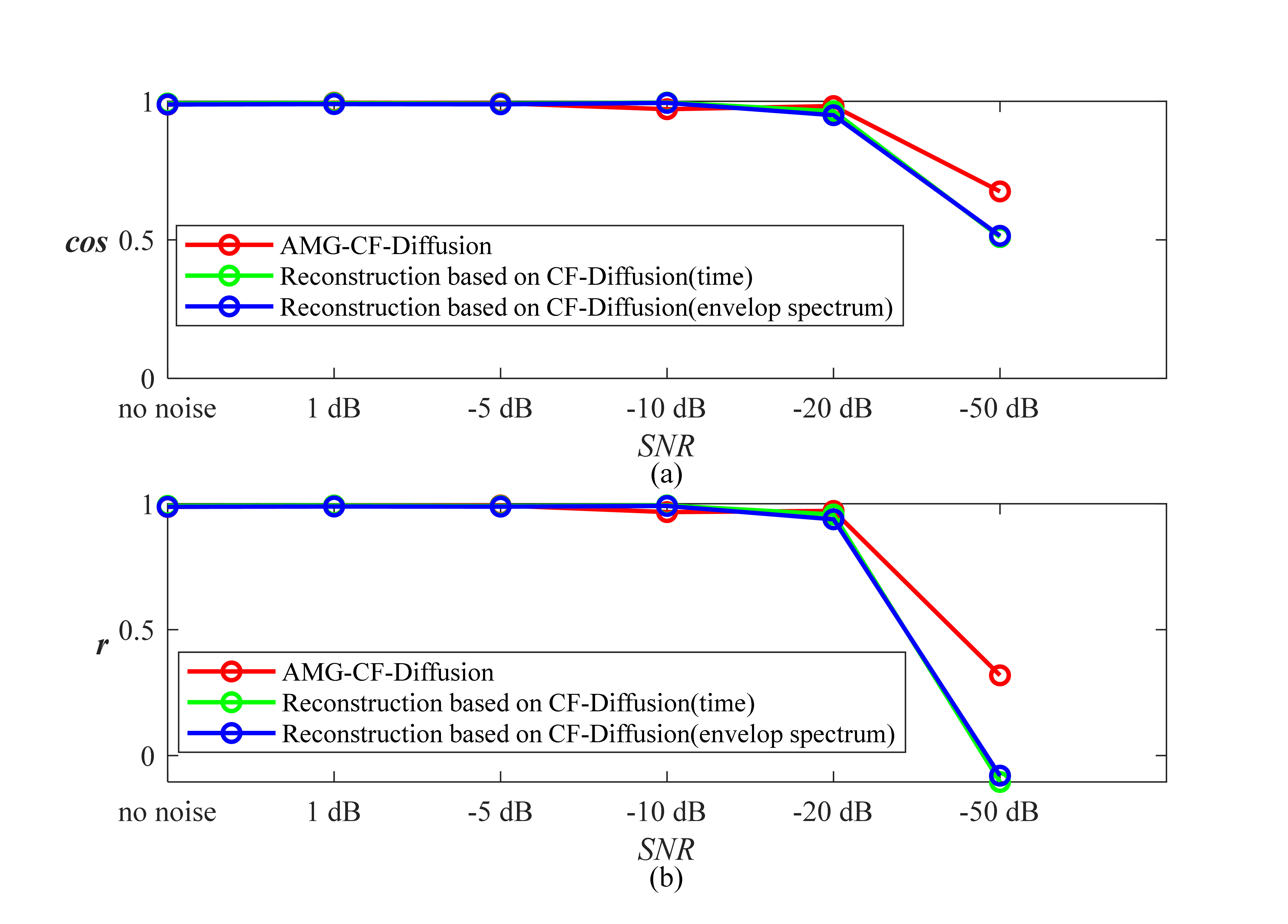}
\caption{The quantitative analysis of similarity between actual degradation and HI curves constructed using different methods based on CF-Diffusion: (a) HI derived from time domain reconstruction, (b) HI derived from envelope spectrum reconstruction .}
\label{fig_simu_ablation}
\end{figure}

\subsection{IMS experiment}
\subsubsection{experimental setting}
\par
\begin{figure}[!t]
\centering
\includegraphics[width=3.0in]{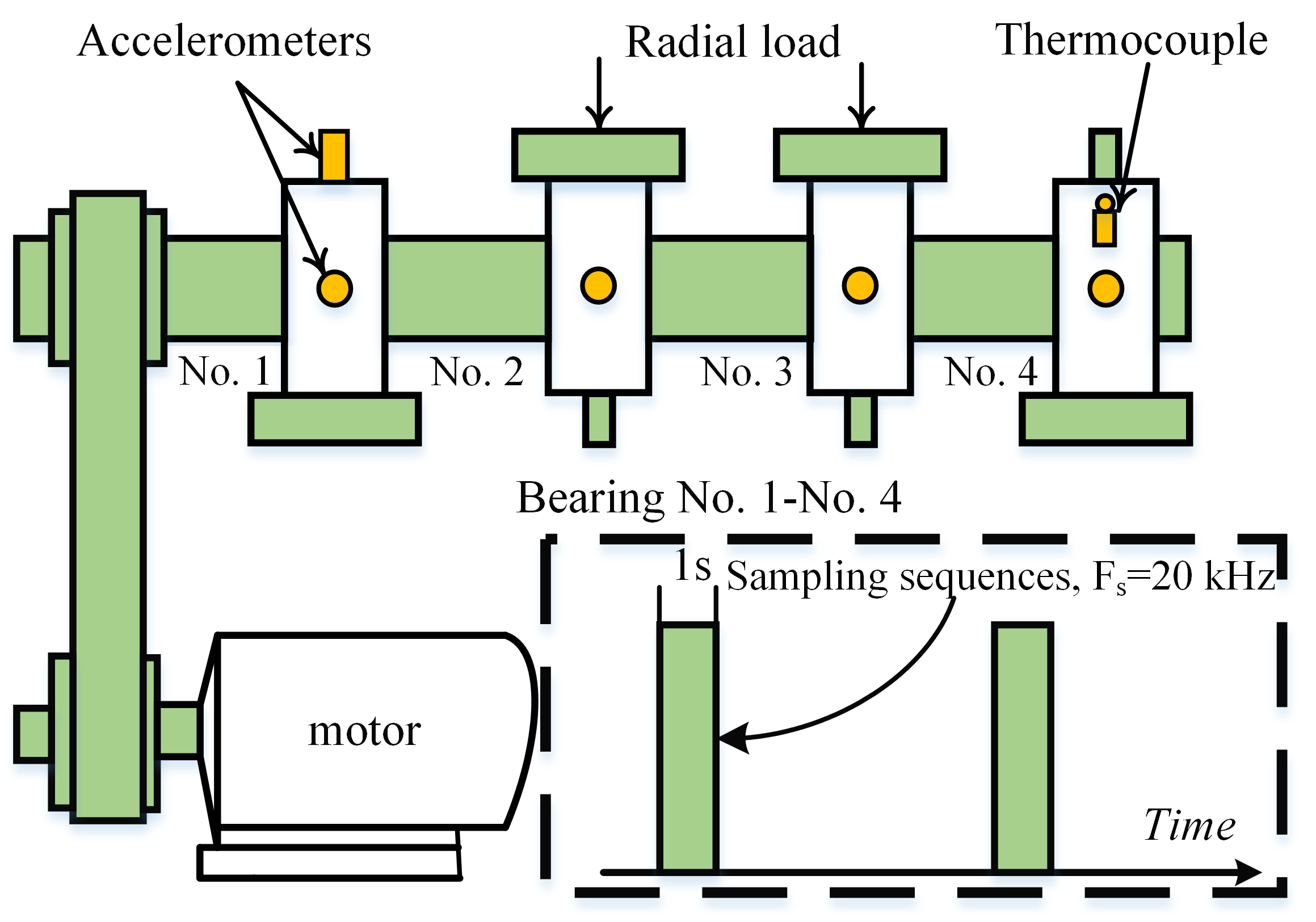}
\caption{The Test rig of the IMS dataset}
\label{fig_test_rig_ims}
\end{figure}

 In this case study, we will use a real dataset to further evaluate the superiority of our proposed method under practical working conditions. We selected the IMS dataset due to its widespread use in the evaluation of numerous health monitoring methods \cite{qiu2006wavelet,hu2023wasserstein,hsu2023comparison}. The test rig and the corresponding sensor placement are illustrated in Fig. \ref{fig_test_rig_ims}. The sampling frequency was set at 20kHz, using Rexnord ZA-2115 bearings. The experiment comprises three sub-datasets, from which the subsets No. 2 and No. 3 were selected due to their identical channel arrangements. We used monitoring data from the bearing No.3 in the dataset No. 3 as the training set, which included healthy samples and a small number of anomalies -- specifically, the first 200 original measured samples as healthy samples and the last 10 original measured samples as anomalies. Furthermore, condition monitoring data from the No. 3 bearing in dataset No. 2 were used for performance evaluation. This dataset consists of 984 original measured samples, with each original measured sample providing 20,480 data points. They can be segmented into 20 samples per original measured sample, each sample being 1,024 data points long. Notably, each bearing in Fig. \ref{fig_test_rig_ims}, was monitored through a single channel. By the test's end, failure in the outer ring of bearing No. 1 was detected in both subset No. 2 and subset No. 3. Specifically, in many practical applications, the measuring points are often positioned at a distance from the location of failure. Therefore, we used the condition monitoring data from bearing No. 3 for performance evaluation to mirror these conditions. 
\par
\begin{figure}[!t]
\centering
\includegraphics[width=3.5in]{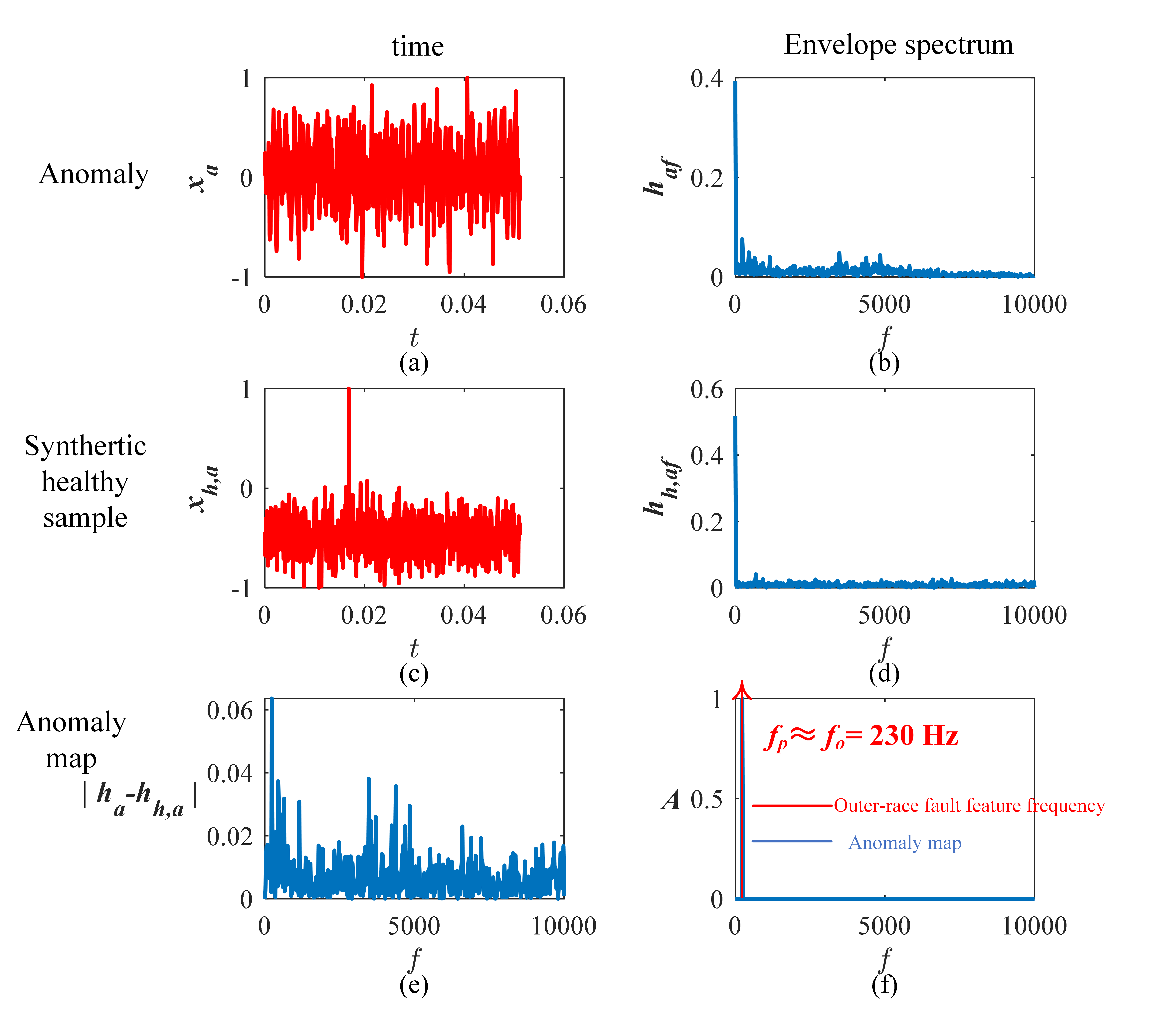}
\caption{The ordinary form of the original anomaly sample and its corresponding synthetic sample and the obtained anomaly maps. }
\label{fig_od_form_ims}
\end{figure}

 In this experiment, the network architecture was largely consistent with that used in Case I, with the exception that the hidden size of the multi-head attention layer was adjusted to 1024, matching the length of each samples. The training parameters remained unchanged from Case I. After training the CF-Diffusion network, it was able to synthesize samples that closely mimicked the distribution of actual healthy samples. Visual representations of the original healthy samples, anomalies, and the synthesized samples are displayed in Fig. \ref{fig_od_form_simu}. Subsequently, an anomaly map $\boldsymbol{A}$ was generated as described in Eq. (\ref{eq_ano_map}) and is illustrated in Fig. \ref{fig_od_form_ims}. Comparison between the anomaly map and the theoretical fault feature frequency of the bearing revealed a pulse corresponding to the outer-race fault, indicating the model's ability to identify specific fault types without pre-labeled fault data. This anomaly map provides actionable insights into the nature of the faults present in the samples, offering a basis for the targeted construction of Health Indicators (HIs). This method not only enhances diagnostic accuracy but also adds an element of explainability to the fault detection and unsupervised fault diagnostics process.
\subsubsection{Results and discussion}
\par
\begin{figure}[!t]
\centering
\includegraphics[width=3.5in]{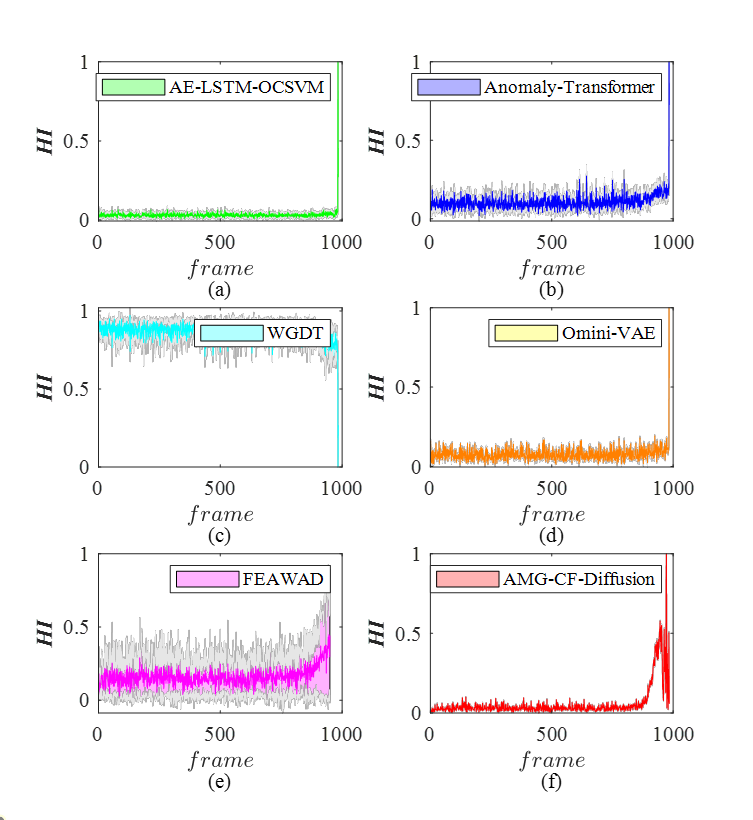}
\caption{The HI curves constructed with the proposed method and different baseline models under different noise interference. (a)AE-LSTM-OCSVM (b) Anomaly-Transformer (c) WGDT (d) Omini-VAE (e) FEAWAD (f) AMG-CF-Diffusion)}
\label{fig_comp}

\end{figure}

 Following the generation of an anomaly map as per Eq. (\ref{eq_ano_map}), we constructed HIs for No. 2 dataset of the IMS dataset, as depicted in Fig.\ref{fig_comp}. To evaluate the effectiveness of our proposed method in health monitoring and early fault detection, we benchmarked it against several established models, including the Anomaly-transformer\cite{xu2021anomaly}, Omni-VAE\cite{su2019robust}, FEAWAD\cite{zhou2021feature}, and WGDT\cite{hu2023wasserstein}. These models were selected as they represent the SOTA unsupervised and weakly-supervised methods. The HIs derived from our method demonstrated less fluctuation under normal operating conditions compared to those from the baseline models, as seen in Fig.\ref{fig_comp}. For a quantitative assessment, we applied standard anomaly detection metrics such as AUROC and AUPRC. In addition, we analyzed early fault detection capabilities using a common threshold selection strategy outlined in \cite{abboud2019advanced} and detailed in Eq. (\ref{eq_thre}). Furthermore, we established the detection results of the AE-LSTM-OCSVM as our baseline for comparison, analyzing how many measured samples earlier the other baseline models and our proposed AMG-CF-Diffusion could detect faults. This approach allows us to present quantitative in a more intuitive manner. These results, detailed in Table. \ref{Table_ims_res} demonstrate that our proposed method achieves the highest scores across these three quantitative metrics (AUROC, AUPRC, Earlier detection time), indicating that it significantly outperforms the other baseline models in both health monitoring and early fault detection. 
\subsubsection{Ablation studies}
\par
\begin{figure}[!t]
\centering
\includegraphics[width=3.5in]{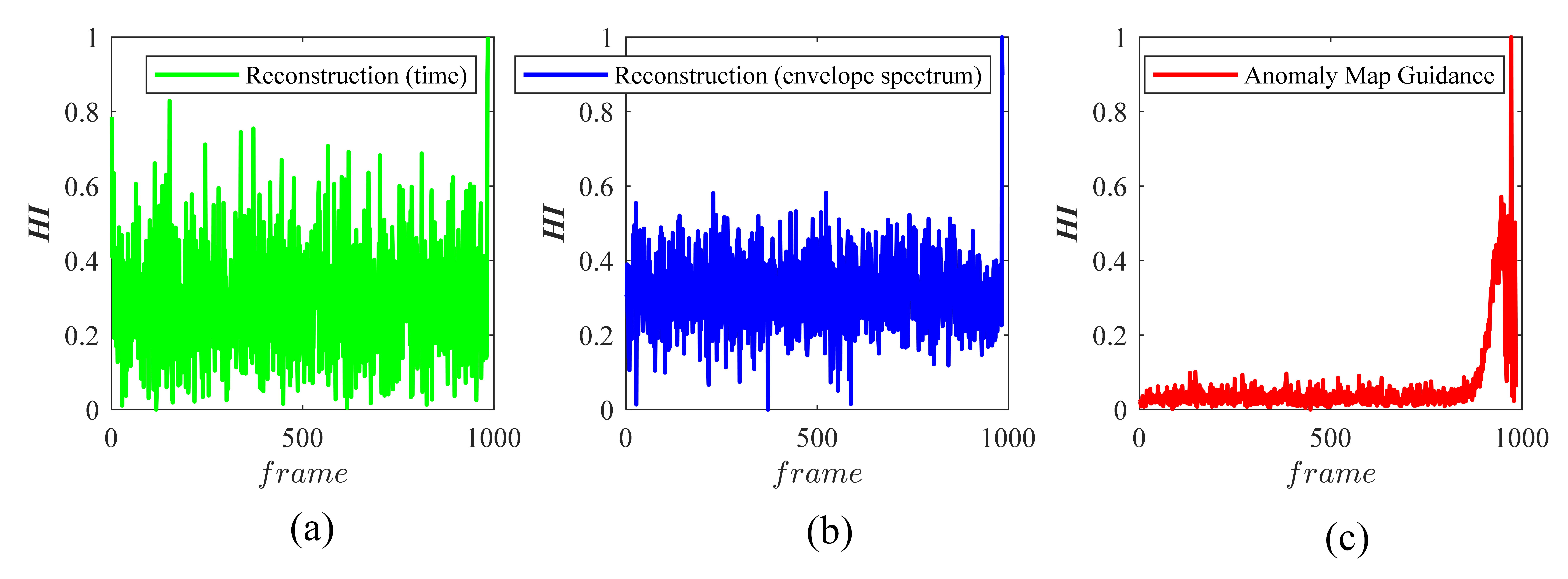}
\caption{The comparative studies of different ways to construct HI. (a) CF-Diffusion (time), (b) CF-Diffusion (envelope spectrum), (c) AMG-CF-Diffusion }
\label{fig_ims_ablation}
\end{figure}

 Furthermore, we conducted ablation studies to assess the effectiveness of the classifier-free mechanism. The experimental approach was consistent with that used in Case I, focusing on performance comparisons of HIs generated using different methods: the time-domain reconstruction based on the CF-Diffusion model, the envelope-spectrum reconstruction based on the CF-Diffusion model, and our anomaly map guidance mechanism, also based on the CF-Diffusion. The results, displayed in Fig. \ref{fig_ims_ablation}, reveal that the HIs derived from time-domain and envelope-spectrum-reconstructions exhibit greater fluctuation compared to those produced using the anomaly map guidance mechanism. Notably, the HI curve from the anomaly map guidance mechanism, illustrated in Fig. \ref{fig_ims_ablation}(c), demonstrates a steady and gradual upward trend that aligns closely with the degradation pattern of the bearing. This outcome emphasizes the superior stability and reliability of the anomaly map guidance mechanism in tracking the progressive deterioration of the machinery.
\begin{equation}
 threshold = \mu_{\boldsymbol{HI}} \pm 6\sigma_{\boldsymbol{HI}}
 \label{eq_thre}
\end{equation}

\par 

\begin{table}[htbp]
\centering
\caption{The quantitative results of different models in health monitoring and early fault detection}
\begin{tabular}{cccc}
\toprule
Method & AUROC & AUPRC & Earlier detection time\\
\midrule
AE-LSTM-OCSVM & 0.60 & 0.17 & 0 \\ 
Anomaly-Transformer & 0.75 & 0.28 & 0 \\
WGDT & 0.76 & 0.30 & 0\\
Omini-VAE & 0.62 & 0.21 & 0 \\
FEAWAD & 0.79 & 0.59 & 40\\
AMG-CF-Diffusion & $\boldsymbol{0.91}$ & $\boldsymbol{0.77}$ & $\boldsymbol{97}$ \\
\bottomrule
\label{Table_ims_res}
\end{tabular}
\end{table}

\section{Conclusion}
\par In this paper, we proposed an anomaly map guidance mechanism for health monitoring and early fault detection of rotating machines, designed to enhance robustness against strong noise interference and provide clearer explanations for fault types associated with anomalies. Our method leverages the CF-Diffusion model to generate controlled healthy samples for each measured sample of condition monitoring data, significantly simplifying the classifier training challenges that the CG-Diffusion model faces when generating time series data. Building on this, an anomaly map is constructed based on the distribution differences between the generated healthy samples and the original samples in the envelope spectrum. This anomaly map not only can identify potential fault types but also supports the construction of HI, minimizing the influence of extraneous noise. Through two case studies, our approach has demonstrated enhanced effectiveness and robustness in various health management related tasks, including health monitoring, degradation tracking, and early fault detection, outperforming several established baseline models.

\section*{CRediT authorship contribution statement}
\par \textbf{Wenyang Hu}: Writing–review \& editing, Writing–original draft, Validation, Software, Methodology, Conceptualization. \textbf{Gaetan Frusque}: Writing–review \& editing, Conceptualization. \textbf{Tianyang Wang}:  Supervision, Resources, Funding acquisition. \textbf{Fulei Chu}: Supervision, Resources, Funding acquisition. \textbf{Olga Fink}: Writing–review \& editing, Supervision, Resources, Conceptualization.



\bibliographystyle{elsarticle-num} 
\bibliography{reference}






\end{document}